\documentclass[letterpaper]{article} 

\usepackage{times}  
\usepackage{helvet}  
\usepackage{courier}  
\usepackage{url}  
\usepackage{graphicx}  
\usepackage[utf8]{inputenc} 
\usepackage[T1]{fontenc}    
\usepackage{booktabs}       
\usepackage{amsfonts}       
\usepackage{nicefrac}       
\usepackage{microtype}      
\usepackage{tikz}
\usetikzlibrary{shapes.misc, positioning}
\usepackage{thmtools,thm-restate}
\usepackage{parskip}
\usepackage{authblk}
\usepackage{amsmath}  
\usepackage{amsthm}
\usepackage[numbers,sort]{natbib}

\usepackage{floatrow}
\newfloatcommand{capbtabbox}{table}[][\FBwidth]

\usepackage{macros}
\usepackage{basic}

\title{Multimodal Data Curation via Object Detection and Filter Ensembles}

 \author[$\dagger$]{Tzu-Heng Huang$^*$}
 \author[$\dagger$]{Changho Shin$^*$}
 \author[$\dagger$]{Sui Jiet Tay}
 \author[$\dagger$]{Dyah Adila}
 \author[$\dagger$]{Frederic Sala}

 \affil[$\dagger$]{University of Wisconsin-Madison}
 \affil[ ]{\footnotesize{\texttt{\{thuang273, cshin23, adila, fredsala\}@wisc.edu}, \footnotesize{\texttt{jiet9000@gmail.com}}}}

\begin{document}

\maketitle

\def\thefootnote{*}\footnotetext{Equal contribution to this work}
\def\thefootnote{\arabic{footnote}}

\begin{abstract}

We propose an approach for curating multimodal data that we used for our entry in the 2023 DataComp competition filtering track\footnote{Appeared in the Workshop of Towards the Next Generation of Computer Vision Datasets (TNGCV) on ICCV 2023.}. 
Our technique combines object detection and weak supervision-based ensembling.
In the first of two steps in our approach, we employ an out-of-the-box zero-shot object detection model to extract granular information and produce a variety of filter designs.
In the second step, we employ weak supervision to ensemble filtering rules. 
This approach results in a 4\% performance improvement when compared to the best-performing baseline, producing the top-ranking position in the small scale track at the time of writing.
Furthermore, in the medium scale track, we achieve a noteworthy 4.2\% improvement over the baseline by simply ensembling existing baselines with weak supervision.
\end{abstract}

%
%
%
%
%
%
\section{Introduction} \label{sec:intro}

%
%



Multimodal models, such as CLIP \cite{radford2021clip}, DALL-E \cite{ramesh2021dalle}, Stable Diffusion \cite{rombach2022stable}, Flamingo \cite{alayrac2022flamingo}, and FLAVA \cite{singh2022flava} have shown unprecedented performance in many vision-language tasks. 
Massive datasets collected from the web play a crucial role in these successes. 
As a result, there is renewed interest in \emph{data-centric} approaches \cite{zha2023data, jarrahi2022principles} to machine learning, focusing on data rather than models, architectures, training approaches, etc.
An important part of this data-centric approach involves community efforts to curate enormous open-source vision-language datasets via large crawls of the web \cite{schuhmann2021laion400m, schuhmann2022laion5b}.

While offering impressive scale, raw web-crawled data can be noisy and lack appropriate selection.
Data curation is therefore crucial. 
However, there are many questions on what might be the right approach for curation in the context of training large-scale models.
In order to shed light on these, the \emph{DataComp} competition\footnote{\url{https://www.datacomp.ai/}} invites users to propose a variety of data curation approaches while fixing model architectures, training procedures, and a raw data pool \cite{gadre2023DataComp}.

In this work, we document our data curation framework and report performance results for the DataComp filtering track at small and medium scale\footnote{\url{https://www.datacomp.ai/leaderboard.html}}. 
Our approach is predominantly based on object detection and filter ensembles. 
In the small scale case, we include various additional rules generated from higher-order granular information via object detection, which yields 4.0\% improvement over the best-performing existing baseline (CLIP score (L/14 30\%)).
Additionally, in the medium scale case, we ensemble baseline filters, which provides 4.2\% improvement over the top-performing baseline (Image-based $\cap$ CLIP score (L/14 30\%)). 
%

\section{Data Curation Framework} \label{sec:methods}

\begin{figure}[t!]
    \includegraphics[width=\linewidth]{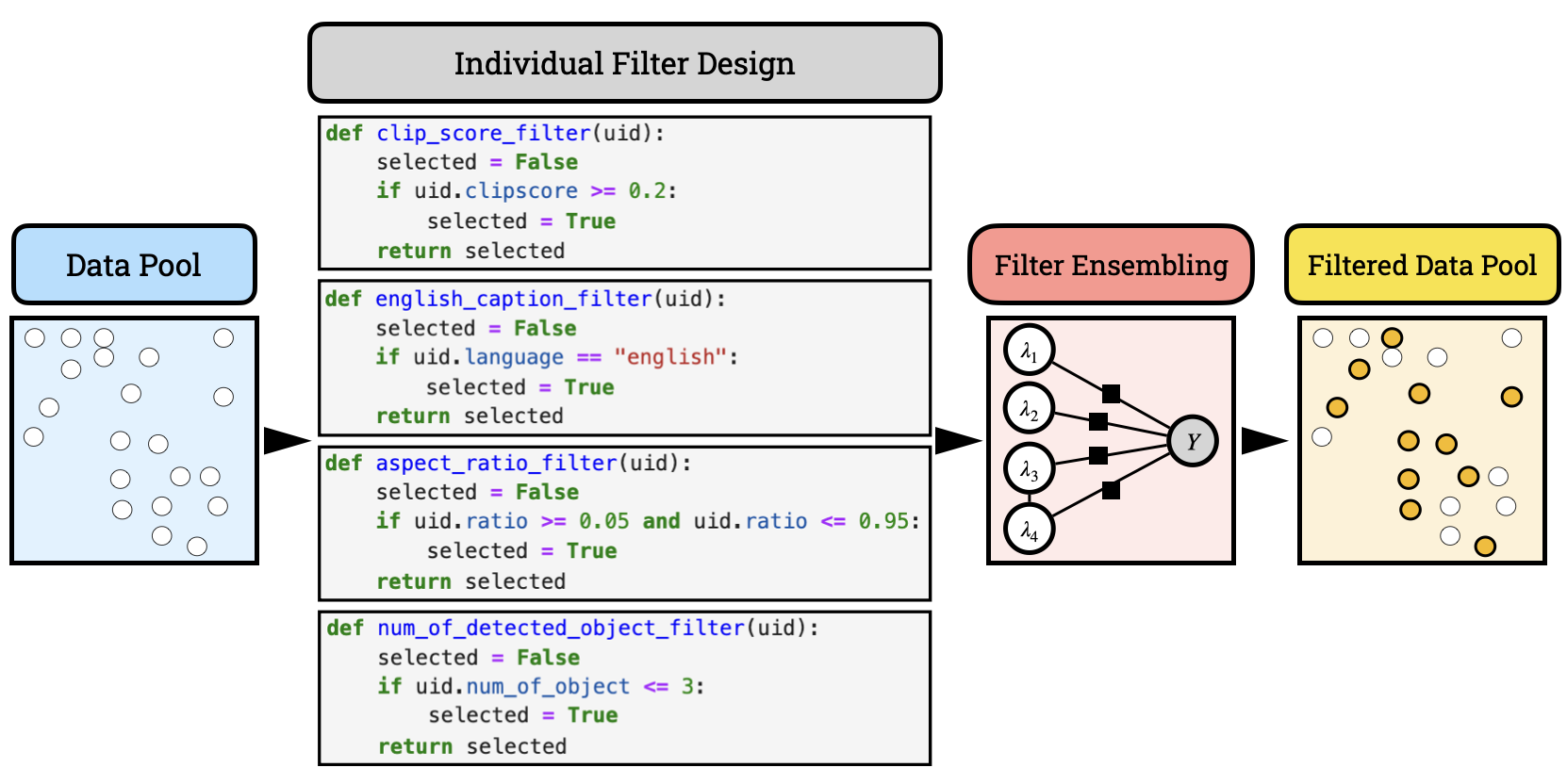}
    \centering
    \vspace{1pt}
    \caption{Overall workflow of the data curation framework. Each data point in the raw data pool is passed to individual filters, which can be designed by human heuristics, pre-computed CLIP scores, or inference results from other off-the-shelf models. We employ Grounding DINO, a zero-shot object detection model, to identify objects mentioned in the image caption. After each designed filter processes, we ensemble filtering results and curate the final refined data pool.}
    \label{fig:framework}
\end{figure}

Our curation approach mainly focuses on filtering and ensembling. 
We provide an overall workflow of the proposed framework in Figure \ref{fig:framework} and discuss each component in-depth.
Broadly, our framework involves two steps. 
First, we design individual filters by considering multiple filtering sources such as existing Datacomp baselines with human heuristics, provided CLIP scores, and an additional component, object detection filters. 
Next, we tune the thresholds in filters to establish refining rules by evaluating the performance with downstream tasks in the Datacomp benchmark.
At last, we select established filters and apply a weak supervision algorithm \cite{Ratner18, shin21universalizing} to ensemble filters and aggregate each filtered result to curate the final dataset.

\subsection{Filtering Method Design}

\begin{figure}[t!]
    \includegraphics[width=\linewidth]{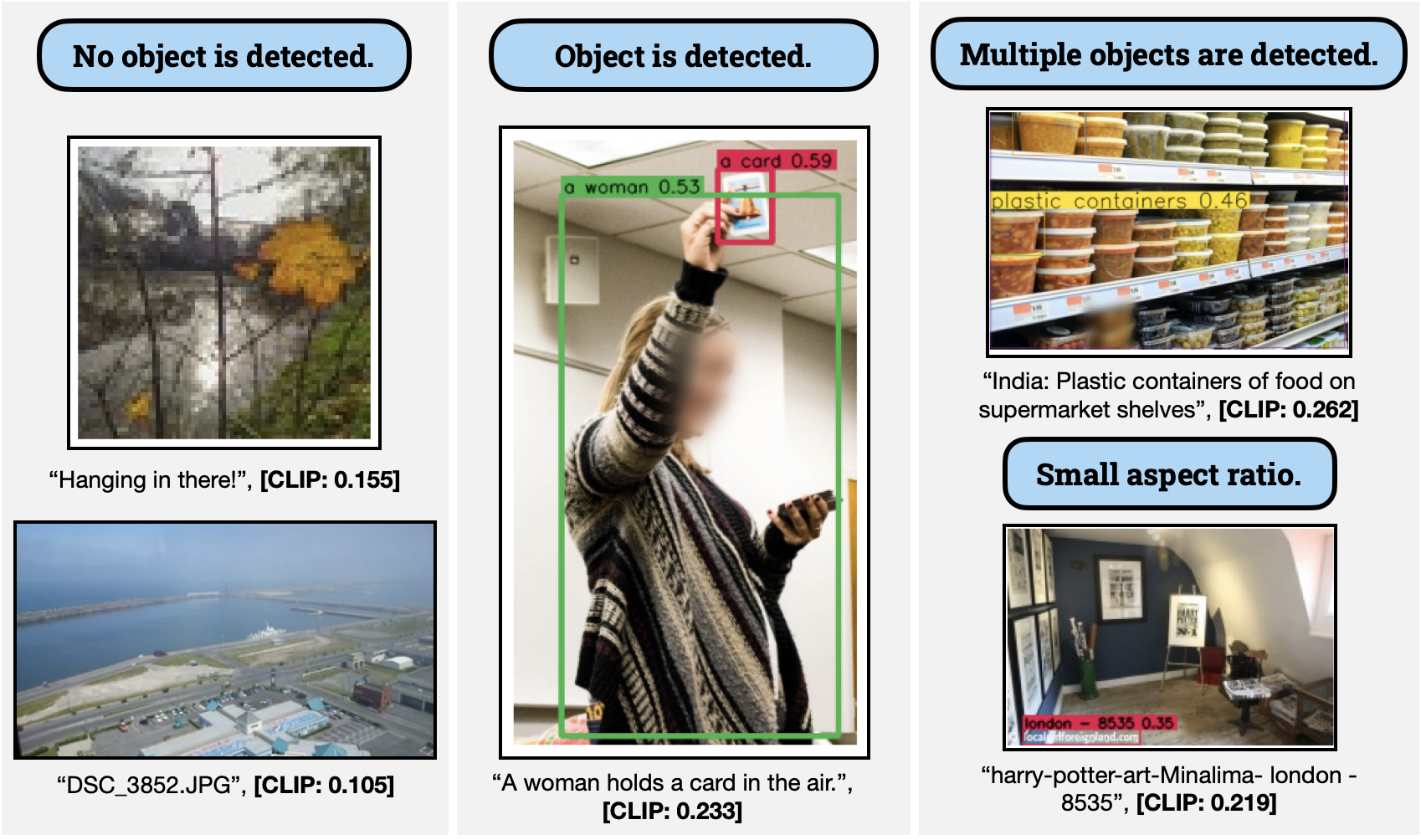}
    \centering
    \vspace{0.5pt}
    \caption{We showcase various image samples with their caption and CLIP score in the small scale dataset and annotate recognized objects through Grounding DINO. Nearly 38\% of the images do not have any identified objects, while 18\% of them have multiple detected objects. Additionally, around 3\% of the images have tiny detected objects. These results offer rich information, which can be used and combined with heuristics to design additional filtering rules.}
    \label{fig:dino examples}
\end{figure}

Our created filters mainly rely on two data pruning approaches and their intersections.
The first approach leverages provided CLIP scores (L/14) to select ``top x\%'' of the images that have high similarity score, while the second approach uses the inference results from a zero-shot object detection model to design rules for refining images.

In this work, we employ Grounding DINO \cite{liu2023grounding}\footnote{https://github.com/IDEA-Research/GroundingDINO} to identify objects mentioned in the image caption and anchor their locations. 
There are three types of inference outcomes provided by Grounding Dino: bounding boxes of detected objects, predicted logit scores for each object (scaled from 0 to 1), and the phrases for each detected object.

Such zero-shot object detection models have several advantages.
First, the inference results offer a certain level of certification that the mentioned objects exist in the image.
In addition to the pre-computed CLIP scores, identification results are more granular, providing information that can be included in filter designs based on interpretable heuristics.
Furthermore, zero-shot object detection models possess the capability to detect objects that are unseen in the predefined set of classes used for training.
This feature makes our filtering framework ideal for recognizing new objects and addressing diverse scenarios in image-text datasets obtained from web crawling on a large scale.
Finally, Grounding DINO is simple and out-of-the-box, which does not require human efforts to customize prompts for candidate objects to query whether exist or not. 
Grounding DINO automatically detects phrases from the given caption and links corresponding visual elements to locate, allowing our filtering method to be easily applied to a massive amount of images.

Several image examples are categorized and displayed in Figure \ref{fig:dino examples}. With the inference results from Grounding DINO, we convert these into \emph{filtering conditions} to refine images in accordance with three types of human heuristics. The conditions are the following:
\begin{enumerate}
    \item \textbf{Predicted logit scores}: the logit score acts as a measure of how certain Grounding DINO is in identifying specific objects. To eliminate images with low logit scores, we take the average and the maximum scores of detected objects within an image and establish a threshold to refine the data pool,
    \item \textbf{Number of detected objects}: Grounding DINO is capable of detecting multiple objects. Since the image captions are brief and short, the objects mentioned in them are expected to fall within a certain range. Therefore, we determine the total number of objects detected in an image and discard any images that have an out-of-range number of objects,
    \item \textbf{Aspect ratio of detected objects}: In order to eliminate object localization that lacks significance, we compute the aspect ratio of every object that is detected and calculate the average ratio present in the image. Afterward, we use a threshold to eliminate images that are either too small or too large in the frame.
\end{enumerate}

Obtaining these filters is cost-effective. They serve the basic purpose of checking the presence of particular objects mentioned in the caption within an image.
Additionally, they are designed to be easily integrated with other contributed filters for greater adaptability and enhancement.
To illustrate this notion, by combining object detection filters with CLIP score filters and analyzing their intersection, we can discard images devoid of any objects but with a high CLIP score.


\subsection{Ensembling Filtering Methods}

%

When multiple results from different filters are available, the most effective approach is to combine them using an ensembling method.
Our ensembling strategy borrows from the rich literature on weak supervision \cite{bach2018snorkel, Ratner19metal, chen2022shoringliger, shin21universalizing}.

Let the raw dataset be $\mathcal{D}=\{x_{i}\}_{i=1}^n$, where $x_{i}$ is an image-text pair, and denote $y \in \{0, 1\}^{n}$ be the inclusion labels processed by a given filter --- keep data point $x_i$ if $y_i=1$ --- such that $D_y=\{x_{i}|y_{i}=1, i=1,\ldots,n\}$ be the refined dataset.
Define the downstream task loss $\mathcal{R}(x, y; \mathcal{A})$, given a training model and algorithm $\mathcal{A}$, and let $y^*=\arg\min\mathcal{R}(x, y; \mathcal{A})$.
While $y^*$ can be obtained by brute force search over $O(2^n)$ combinations, this is  not practically possible even for the small scale case in DataComp, where $n=12.8$ millions.
Instead, we use a variety of filtering methods $\lambda^j$ $(j=1,\ldots,m)$ such that $\lambda^j(x)\in \{0, 1\}$.
These designed filters are assumed to have some level of accuracy with respect to $y^*$.

Hence, given $m$ filters, the most basic ensemble approach is majority voting, i.e.
\[\hat{y}_i=\frac{1}{m}\sum_{j=1}^m \lambda^{j}(x_i)\]
However, majority voting fails to consider the accuracy and correlation of filtering methods.
A standard approach in weak supervision is to encode filters' accuracy and correlations with the Ising model \cite{bach2018snorkel, Ratner19metal, fu2020triplet}, i.e.,
\[
p(\lambda^1, \ldots, \lambda^m, x_i, y_i)= \frac{1}{Z} \exp \left( \sum_{j=1}^{m} \theta_{j}\lambda^j(x_i)y_i + 
\sum_{(j,k)\in E}\theta_{j,k}\lambda^{j}(x_i)\lambda^{k}(x_i) + \theta_{Y}y_i \right)
\]
,where $\theta_{j}, \theta_{j,k}, \theta_{Y}$ are the canonical parameters encoding accuracy, correlation, and class balance respectively. 
$E$ is the set of candidate filters correlations, and $Z$ is a normalization constant to ensure the probability is valid.
After learning the parameters in the Ising model, we can infer the most probable inclusion labels by computing 
\[\hat{y}_i = \arg\max_{y'\in \{0, 1\}}
p\left(y_i=y'| \lambda^1(x_i), \ldots, \lambda^m(x_i)\right)\]
Finally, $\hat{y}_i$ is the aggregated decision to include $x_i$ or not, and $D_{\hat{y}}=\{x_{i}|\hat{y}_{i}=1, i=1,\ldots,n\}$ becomes the final curated dataset to be used to train the model.
\section{Experiments} \label{sec:exps}

\paragraph{Implementation Details}
We document our implementation and report each step in detail. 
We used the img2dataset package\footnote{https://github.com/rom1504/img2dataset} to download small and medium scale datasets in the filtering track, succeeding in downloading 11.9M (93.32\%) samples and 115.0M (89.89\%) samples. 
In the small scale case, we double-checked that the amount of downloaded data is comparable to the DataComp team's data by reproducing ``Baseline: CLIP score (L/14 30\%)'', which gives a slightly lower but comparable performance (ImageNet accuracy 0.045/0.051 and Average performance 0.168/0.173). 

In Grounding DINO, the inference rate is not consistent and can be slow when processing with the original image size. 
To improve computational efficiency, we resized the images. 
However, this trade-off between performance and inference rate must be taken into account.
After attempting to resize with various dimensions including (224, 224), (400, 400), and (800, 800), we ultimately settled on using (400, 400) as it provided similar performance.
Among variants of Grounding DINO, we choose Grounding-DINO-T, which used Swin-T \cite{liu2021swin} as an image backbone and BERT \cite{kenton2019bert} as a text backbone.

In the ensembling step, we employed Snorkel \cite{bach2018snorkel}\footnote{https://github.com/snorkel-team/snorkel.}, a well-known framework for weak supervision.
We set the class balance parameter to 0.3 and 0.2 for small and medium scale datasets respectively, based on prior works \cite{gadre2023DataComp} that investigated the relationship between the size of the filtered dataset and downstream task performance.
To ensemble multiple filter results, we trained the Snorkel label model for 1000 epochs with learning rate 0.01. All the experiments were performed on an NVIDIA A6000 GPU. 

\paragraph{Filtering Conditions}
Next, we discuss considered conditions when designing the three types of object detection filters --- logit score, detected number, and aspect ratio.
First, if there is no object identified by Grounding DINO, the output is empty. 
There are about 38\% of images in this setting, and we eliminated them from the raw data pool.

For the logit score filter, we selected image examples with the highest average score and highest maximum score based on the top 30\%.
Another filter was designed to detect numbers within a specific range of 1 to 4 and 1 to 3.
The aspect ratio filter discarded images with an average aspect ratio smaller than 5\% or larger than 95\%.
Finally, to ensure well-aligned image-text examples, we intersected each of the above filters with various CLIP score filters, including CLIP L/14 30\%, 50\%, and 55\%.
Our design of these thresholds took into account the size of the resulting dataset.
\section{Results}

\begin{table*} [t]
\caption{Performance comparison of individual filters in small scale dataset.}
   \label{tab:small_individual_results}
   \centering
   \begin{tabular}{llll}
     \toprule
    Method & Dataset size & ImageNet acc. & Average perf.\\
    \midrule
    CLIP L/14 30\% (DataComp) & 3.84M & .051 & .173 \\
    \midrule
    OD Avg. Logit 30\% $\cap$ CLIP L/14 50\% &  3.84M & .054 & .164 \\
    OD Avg. Logit 30\% $\cap$ CLIP L/14 30\% &  2.38M & \textbf{.059} & \textbf{.172} \\
    \midrule
    OD Max. Logit 30\% $\cap$ CLIP L/14 30\% &  2.39M & .054 & .172 \\
    OD Max. Logit 30\% $\cap$ CLIP L/14 50\% &  3.84M & \textbf{.059} & \textbf{.173} \\
    \midrule
    OD Num. of Objects ($\leq 4$) $\cap$ CLIP L/14 50\% &  4.43M & .052 & .170 \\
    OD Num. of Objects ($\leq 3$) $\cap$ CLIP L/14 55\% &  4.61M & .050 & .173 \\
    OD Num. of Objects ($\leq 3$) $\cap$ CLIP L/14 50\% &  4.36M & \textbf{.052} & \textbf{.174} \\
    \midrule
    OD Avg. Aspect Ratio $\cap$ CLIP L/14 55\% &  4.19M & .053 & .173 \\
    OD Avg. Aspect Ratio $\cap$ CLIP L/14 50\% &  3.98M & \textbf{.053} & \textbf{.178} \\
    \bottomrule
   \end{tabular}
 \end{table*}

\begin{table*} \label{sec:ensemble_comparison}
\caption{Performance comparison of ensemble filters in small scale dataset.}
   \label{tab:small_ensemble_results}
   \centering
   \begin{tabular}{llll}
     \toprule
    Method &  Dataset size & ImageNet acc. & Average perf.\\
    \midrule
    MV (baselines) & 2.39M & .060 & .168 \\
    WS (baselines, class balance 0.5) & 6.38M & .043 & .153 \\
    WS (baselines, class balance 0.2) & 2.49M & .058 & .174 \\
    WS (baselines, class balance 0.3) & 3.20M & \textbf{.059} & \textbf{.175} \\
    \midrule
    WS (baselines + All the OD Filters) & 4.10M & .055 & .169 \\
    WS (baselines + OD Max. Logit + OD Avg. Aspect Ratio) & 3.92M & .059 & .172 \\
    WS (baselines + OD Num of Objects + OD Avg. Aspect Ratio) & 4.14M & .056 & .173 \\
    WS (baselines + OD Avg. Logit + OD Max. Logit) & 4.11M & \textbf{.056} & \textbf{.180} \\
    \bottomrule
   \end{tabular}
 \end{table*}

 
\paragraph{Small Scale Dataset} 
In order to evaluate the effectiveness of the proposed object detection filters and tune threshold parameters to optimize results, we examined the performance of each filter on the small scale dataset.
We anticipate that Grounding DINO's outputs will provide additional granular information that can be used to create better filtering rules.
Table \ref {tab:small_individual_results} shows the performance of the curated training dataset with various designed filter combinations.
As expected, when combined with the CLIP score filter, the object detection filters outperform most of DataComp's existing baselines. 
Additionally, by setting the threshold correctly, we find that the aspect ratio filter produces the best average performance of 0.178 among all methods, while another designed filter that counts the number of detected objects achieves the suboptimal performance at 0.174.
With the optimal threshold and its resulting filters, we were able to establish filtering conditions and used them in the ensembling step.

Subsequently, to evaluate the performance of our ensemble method, we began by aggregating the filtered datasets created by DataComp's baselines.
The performance of the final curated datasets, obtained through different filter combinations, is presented in Table \ref{tab:small_ensemble_results}.
We observe that both majority voting and weak supervision techniques yield results that are comparable or even superior to the best individual rule when the dataset is curated solely by the baseline ensemble.
Additionally, weak supervision demonstrates superior performance compared to majority voting by 4.1\%.
These demonstrate the advantages of our ensemble technique and validate the curation framework effectively.

Afterward, we apply this setup and incorporate object detection filters and weak supervision techniques with the class balance parameter 0.3 to ensemble filters for performance enhancement.
Finally, by combining various sources of filters, the combination of baselines and logit score filters achieves the best 4.0\% improvement over the existing best-performing baseline (CLIP L/14 30\%) on average.

\vspace{-8pt}
\paragraph{Medium Scale Dataset}
Our team predominantly focused on the small scale track. However, the ensemble approach with the provided baseline filters can be used to curate the dataset in the medium case without significant additional costs. 
By setting the class balance at 0.2 and using weak supervision techniques, the final refined dataset achieves ImageNet accuracy of 0.305 and average performance of 0.342, exhibiting 4.2\% improvement compared to the top-performing baseline filter.
Our ensemble approach has shown its adaptability by successfully incorporating other contributed filters effectively.
\section{Conclusion}

Large-scale web-crawled data has been widely collected for training multimodal models, producing high-performance models. 
However, web-crawled data often contains noise, low-quality samples, and suffers from poor selection. 
To address this issue, we proposed a framework for refining data pool and improving curated datasets in the 2023 DataComp competition.
Our approach involves designing various filters using off-the-shelf zero-shot object detection models and applying weak supervision-based ensembling techniques.
Through empirical validation, we showed the effectiveness of the designed filters and ensemble methods in our data curation framework on both small and medium scale datasets.
Our approach has resulted in a 4\% performance improvement compared to the best existing baselines, producing the top position in the leaderboard for the small scale track at the time of writing.

\bibliography{reference}
\bibliographystyle{unsrt}

\newpage
\appendix

The appendix is organized as follows. 
In the Appendix \ref{appendix:ensemble_analysis}, we provide an analysis of correlation across all the individual filters we designed and their estimated accuracies. 
Next, we discuss related works in our framework and provide more insights about the properties of contrastive loss in the Appendix \ref{related_work}. 
\section{Correlation and Estimated Accuracy Across Designed Filters} \label{appendix:ensemble_analysis} 

\begin{figure*}[t!]
     \includegraphics[width=\linewidth]{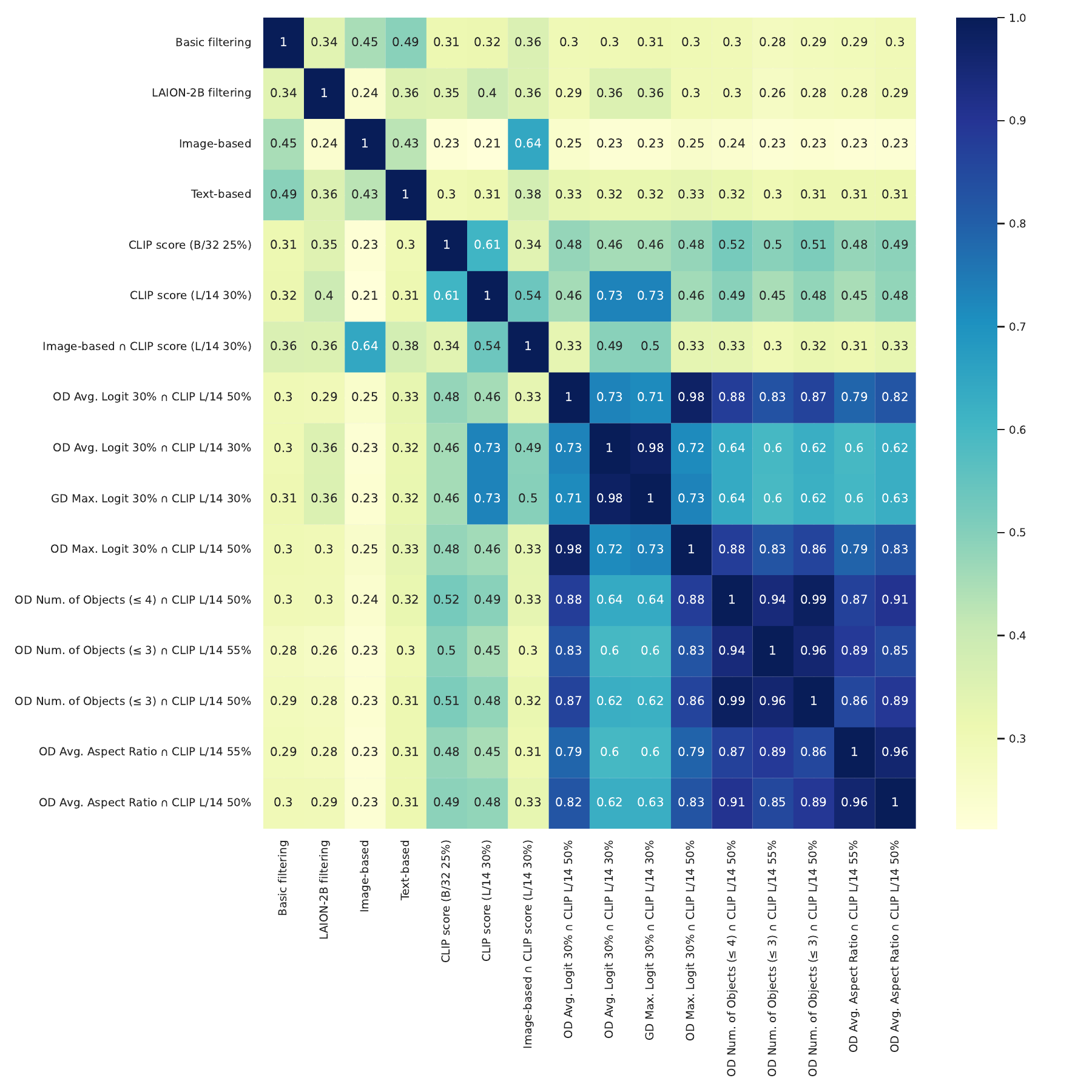}
     \centering
     \caption{Correlation across designed filters. }
     \label{fig:filter_correlation}
\end{figure*}

The correlation between baselines and designed filters listed in Table \ref{tab:small_individual_results} is presented in Figure \ref{fig:filter_correlation}.
As expected, the correlation is low for baseline filters and moderate to high for CLIP and Grounding DINO filters.
This observation could explain the results in Table \ref{tab:small_ensemble_results}, where the baseline ensemble performs well, while including Grounding DINO does not produce the best results due to the high correlation. 
To resolve this issue, the dependency graph can be taken into account, which was avoided for simplicity.

\begin{figure*}[t!]
    \includegraphics[width=0.9\linewidth]{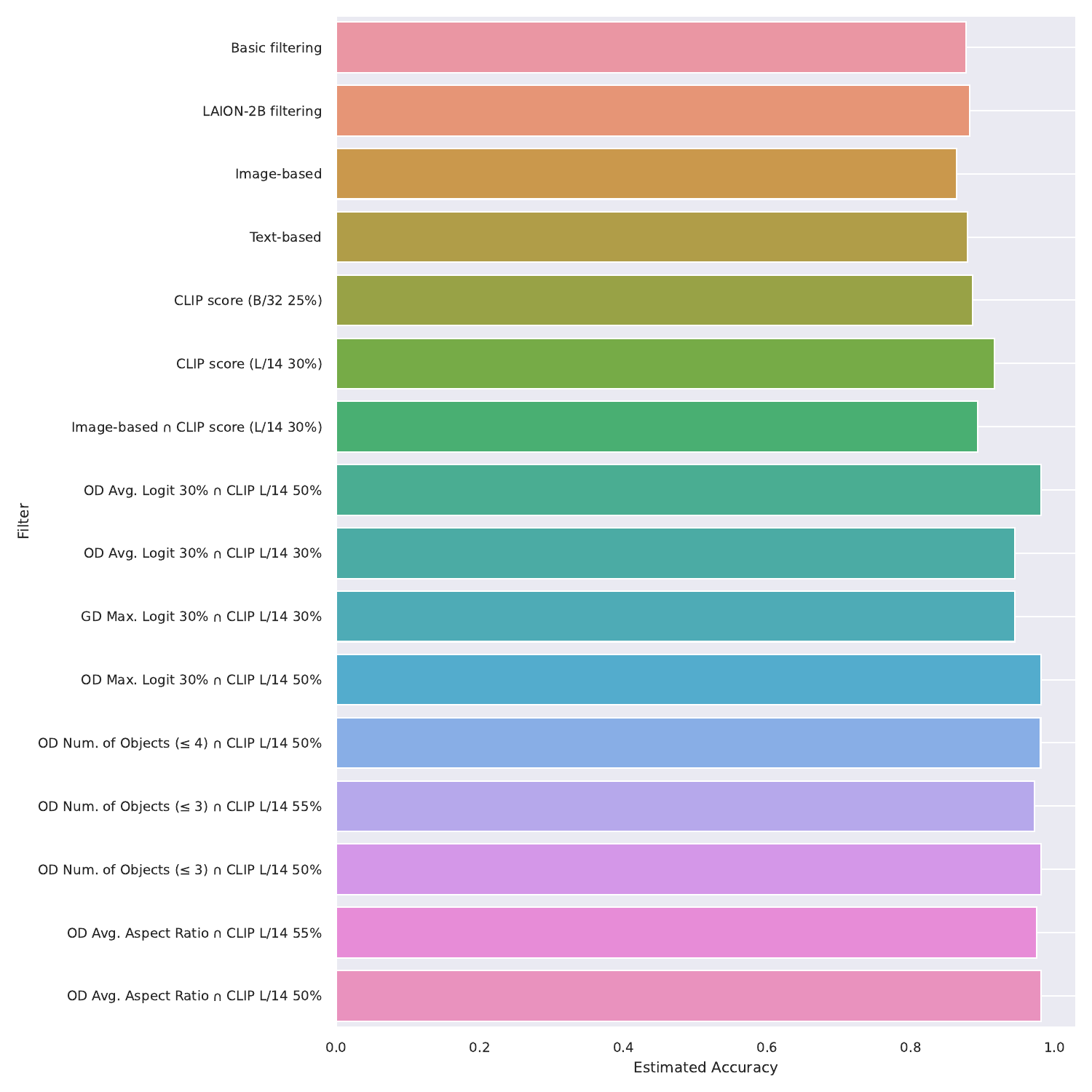}
    \centering
    \caption{Estimated accuracy across designed filters. }
    \label{fig:filter_accuracy}
\end{figure*}

Figure \ref{fig:filter_accuracy} shows the estimated accuracy of each filter in the Ising model (i.e. $P(y=\lambda^{j}(x))$).
Though the estimated accuracy can be affected by the violation of conditional independence, we notice that the estimated accuracy is correlated to the performance that we display in Table \ref{tab:small_individual_results}.

\section{Related Works} 
\label{related_work}


\paragraph{Zero-Shot Object Detection} Zero-shot object detection differs from traditional object detection methods \cite{carion2020end, redmon2016you, girshick2015fast}. Unlike the latter, it is not limited to detecting only pre-defined object classes in the training data. This technique does not require fine-tuning of model parameters to introduce novel object classes. Instead, it uses multi-modal representations and language generalization to perform detection for such objects. Most of the current zero-shot object detection models \cite{minderer2205simple, liu2023grounding, zareian2021open, gu2021open} use CLIP \cite{radford2021clip} as their query module to align textual embeddings and visual components. In our work, we use Grounding DINO \cite{liu2023grounding} as a detector to check that the mentioned objects exist in the image. We then use the returned information to design additional filters.

\paragraph{Weak Supervision} In our ensemble step, we used the most standard label model built on the Ising model \cite{bach2018snorkel, Ratner19metal, fu2020triplet}. While simple, the main drawback of such models is that they do not exploit input geometry, assuming globally uniform accuracy for each filter rule. To overcome such limitations, several existing works incorporate the input space into the label model. \cite{chen2022shoringliger} suggested a partition-based label model, which separated parameters in each input partition. \cite{zhang2023leveraging} applied a Gaussian process and a Bayesian label model to leverage input features. \cite{shin2023mitigating} provided a label model based on accuracy center and slope. While we mainly used Ising model-based techniques to ensemble filters, applying an embedding-based weak supervision approach may be more useful to aggregate filtered results, enabling each filter's strengths in specific input space to be better exploited. 

\paragraph{Properties of Contrastive Loss}
The contrastive loss underpins the training approach of CLIP. 
Though it is natural to filter the data via CLIP scores, some properties of the contrastive loss can provide further insights to curate a dataset.
For example, \cite{chen2021intriguing} found 1) the contrastive loss is feasible with multiple objects while too many objects may undermine model learning, 2) the presence of dominant objects may suppress the learning of feature of small objects, 3) easy-to-learn features may suppress the learning of other features.

The first and the second points support our motivation when we were considering additional components --- object detection filters, especially designing filters considering the number of objects and object relative size.
The third point is related to another work, T-MARS \cite{maini2023tmars}, which enhances the filtering scheme by masking recognized text in the image and then re-scoring as text features are typically easy-to-learn, undermining other features.
As such, exploring the properties of the contrastive loss may yield more insights and heuristics to design and craft filtering methods.

\end{document}